\documentclass[12pt,journal,onecolumn]{IEEEtran}
\IEEEoverridecommandlockouts

\usepackage{cite}
\usepackage{amsmath,amssymb,amsfonts}
\usepackage{graphicx}
\usepackage{textcomp}

\RequirePackage{fix-cm}
\usepackage{amssymb}
\usepackage{graphicx}
\usepackage{url}
\usepackage{amsmath}
\usepackage{color}
\usepackage{epsfig}
\usepackage{amsfonts}
\usepackage{latexsym}
\usepackage{mathtools}
\usepackage{multirow}
\usepackage{algorithm}
\usepackage{algcompatible}
\usepackage[noend]{algpseudocode}
\algnewcommand{\LeftComment}[1]{\Statex \(\triangleright\) #1}

\makeatletter
\def\BState{\State\hskip-\ALG@thistlm}
\makeatother

\usepackage{tikz-qtree}
\usepackage[font=small,labelfont=bf]{caption}
\usepackage[footnote]{acronym}
\usepackage{lscape}
\usepackage{url}
\usepackage{extarrows}
\usepackage[percent]{overpic}
\usepackage{float}
\usepackage{soul}

\newcommand{\bqs}{\begin{eqnarray*}}
\newcommand{\eqs}{\end{eqnarray*}}
\newcommand{\bq}{\begin{eqnarray}}
\newcommand{\eq}{\end{eqnarray}}
\newcommand{\biz}{\begin{itemize}}
\newcommand{\eiz}{\end{itemize}}
\newcommand{\be}{\begin{enumerate}}
\newcommand{\ee}{\end{enumerate}}
\newcommand{\bs}{\boldsymbol}

\newcommand{\eqdef}{\xlongequal{\text{def}}}

\usepackage{ulem}
\usepackage{lettrine}

\begin{document}

\title{Sparsification of the Alignment Path Search Space in Dynamic Time Warping}

\author{
\IEEEauthorblockN{Saeid Soheily-Khah, Pierre-Fran\c cois Marteau,~\IEEEmembership{Member,~IEEE,}}
\IEEEauthorblockA{\textit{} \\
\textit{\small {IRISA, CNRS, Universit\'e Bretagne-Sud, Campus de Tohannic, Vannes, France}} \\
}
}

\maketitle

\begin{abstract}
Temporal data are naturally everywhere, especially in the digital era that sees the advent of big data and internet of things.
One major challenge that arises during temporal data analysis and mining is the comparison of time series or sequences, which requires to determine a proper distance or (dis)similarity measure.
In this context, the Dynamic Time Warping (DTW) has enjoyed success in many domains, due to its 'temporal elasticity', a property particularly useful when matching temporal data. Unfortunately this dissimilarity measure suffers from a quadratic computational cost, which prohibits its use for large scale applications. This work addresses the sparsification of the alignment path search space for DTW-like measures, essentially to lower their computational cost without loosing on the quality of the measure.

As a result of our sparsification approach, two new (dis)similarity measures, namely SP-DTW (Sparsified-Paths search space DTW) and its kernelization SP-$\mathcal{K}_{\mbox{rdtw}}$ (Sparsified-Paths search space $\mathcal{K}_{\mbox{rdtw}}$ kernel) are proposed for time series comparison.  
A wide range of public datasets is used to evaluate the efficiency (estimated in term of speed-up ratio and classification accuracy) of the proposed (dis)similarity measures on the 1-Nearest Neighbor (1-NN) and the Support Vector Machine (SVM) classification algorithms. Our experiment shows that our proposed measures provide a significant speed-up without loosing on accuracy. Furthermore, at the cost of a slight reduction of the speedup they significantly outperform on the accuracy criteria the old but well known Sakoe-Chiba approach that reduces the DTW path search space using a symmetric corridor.
\end{abstract}


\section{Introduction}
\lettrine{D}{ue} to rapid increase in data size, the need for efficient temporal data mining has become more crucial and time consuming during the past decade.
Temporal data naturally appears in many emerging applications such as sensor networks, network security, medical data, e-marketing, dynamic social networks, human mobility, internet of things, etc.   
They can be time series or sequences of time stamped events or even purely sequential data such as strings or DNA sequences.
In general, temporal data refers to data characterized by changes over time and the presence, unlike static data, of temporal dependency among them. 
Therefore the appropriate processing for data dependency or correlation analysis becomes vital in all temporal data processing.

This is in part why temporal data  mining has recently attracted  considerable attention in the data mining community  \cite{Bradley98refininginitial,Halkidi01onclustering,Kalpakis:2001:DME:645496.657890,Keogh:2003:CTS:951949.952156}.  
Typically temporal data mining is concerned with the analysis of temporal data for finding temporal patterns and regularities in sets of temporal data, to define appropriate (dis)similarity measures with temporal alignment capabilities, to extract "hidden" (temporal) structures, classify temporal objects, predict a future horizon and so on.
Since it brings together techniques from different fields such as statistics, machine learning and databases, the literature is diffused among many different sources. 

One main problem that arises during the temporal analysis is the 
 choice or the design of a proper and accurate distance or (dis)similarity measure between time series (or sequences).
There has been a very active research on how to properly define the concept of  "distance", the "similarity" or the "dissimilarity" between time series and sequences. 
Sometimes two time series (or sequences) sharing the same configuration are considered close and their appearances are similar in terms of form, even if they have very different values. However  the value-based proximity measures are the most studied approaches to compare the time series (or sequences). 
Thus,  a crucial issue is establishing what we mean by "similar" or "dissimilar"  series due to their dynamic characters (with or without considering delay or temporal distorsion).

Different approaches to define the distance and the (dis)similarity between time series (or sequences) have been proposed in the literature \cite{MOECKEL1997187,Gunopulos2004,marteau:TPAMI2009,conf/sdm/BatistaWK11,cassisi2012similarity,DBLP:journals/corr/SerraA14,Soheily-Khah2015,li2016scalable}, and for a recent review \cite{Bagnall2017}. 
Among these works, Dynamic Time Warping (DTW) \cite{Bellman,ita75,Kruskall} has been largely  popularized, studied and used in numerous applications. However, one of the main disadvantage of this measure is its quadratic computing cost that should be compared to the linear complexity of the Euclidean distance that can handle any feature sets containing also temporal or sequential features such as shapelets for instance.

The main contribution of this work is to propose a new approach to speed-up the computation of DTW, based on the sparsification of the admissible alignment path search space. Our contribution results in the description of the sparsification method and the proposal for two new (dis)similarity measures, namely SP-DTW (Sparsified-Paths search space  DTW) and its kernelization SP-$\mathcal{K}_{\mbox{rdtw}}$ (Sparsified-Paths search space  $\mathcal{K}_{\mbox{rdtw}}$ kernel). A detailed experimentation is proposed to assess our contribution against the state of the art approaches, dedicated to speed-up elastic distance computation.

The remainder of the paper is organized as follows: in the next section,  a short overview of well-established (dis)similarity measures
are presented, including time elastic measures. Next, in Section 3, we introduce the definition of the proposed sparsification of the alignment path search space that characterized time elastic measures.
Then we introduce two new (sparsified path search space) measures that exploit this sparsification technique.
Later, in Section 3, the conducted comparative experiments and results are discussed. Lastly, Section 4 concludes the paper. 


\section{Times series basic measures}
\label{works}
Mining and comparing time series address a large range of challenges, among them: the meaningfulness of the distance and (dis)similarity measure. 
This section is dedicated to briefly describe the major well-known measures in a nutshell, which have been grouped into two categories:  behavior-based  and value-based measures \cite{soheilykhah:tel-01394280}.

\subsection{Behavior-based measures}
In this category,
time series are compared according to their behaviors (or shapes).
That is the case when time series of a same class exhibit similar behavior (or shapes), and time series of different classes have different behaviors (or shapes). 
In this context, we should define which time series are more similar to each other and which ones are different. 
The Main techniques to recover time series behaviors are: slopes and derivatives comparison,
ranks comparison,  Pearson and temporal correlation coefficients,  and difference between auto-correlation operators. Here, we briefly describe some well-used behavior-based measures.

\subsubsection{Pearson CORRelation coefficient (CORR)}
Let $\textbf{X}=\{\textbf{x}_1,..., \textbf{x}_N\}$  be a  set of time series  $\textbf{x}_i=(x_{i1},...,x_{iT})$ $,i \in\{1,...,N\}$. The correlation coefficient between the time series $\textbf{x}_i$ and $\textbf{x}_j$ is defined by:

\begin{eqnarray}
	\textbf{{\sc corr}}(\textbf{x}_i,\textbf{x}_j) = \frac{\displaystyle\sum_{t=1}^T (x_{i_t} - \overline{\textbf{x}}_i) (x_{j_t} - \overline{\textbf{x}}_j)}{\sqrt{\displaystyle\sum_{t=1}^T {(x_{i_t} - \overline{\textbf{x}}_i)}^2} \ \ \sqrt{\displaystyle\sum_{t=1}^T {(x_{j_t} - \overline{\textbf{x}}_j)}^2}}
\end{eqnarray} 
and  
\begin{center}
$\overline{\textbf{x}}_i = \frac{1}{T} \displaystyle  \sum_{t=1}^T x_{i_t}$
\end{center}

Correlation coefficient was first introduced by Bravais and later, Pearson in \cite{Pearson1896} illustrated that it is the best possible correlation between two time series. Up until now, many applications in different domains such as  speech recognition, system design control, functional MRI and gene expression analysis have used the Pearson correlation coefficient as a behavior proximity measure between the time series (and sequences) \cite{Maca,Ernst,Abra,Cab,Rydell}.  
The Pearson correlation coefficient changes between $\--1$ and $+1$. The case CORR$= +1$, called perfect positive correlation, occurs when two time series perfectly coincide, and the case CORR$ = \--1$, called the perfect negative correlation, occurs when they behave completely opposite. CORR$ = 0$ shows that the time series have different behavior. Observe that the higher correlation coefficient does not conclude the similar dynamics.

\subsubsection{Difference Auto-Correlation Operators (DACO)}
\label{subsec:daco}
The Auto-correlation is a representation of the degree of similarity which measures the dependency  between a time series and a shifted version of itself over successive time intervals.
%
%
%
The Difference Auto-Correlation Operators (DACO) between the two time series $\textbf{x}_i$ and $\textbf{x}_j$ defined by \cite{gaidon2011time}:\\
\begin{eqnarray}
	\textbf{{\sc daco}}(\textbf{x}_i,\textbf{x}_j) = {\Vert  \bf{\tilde{x}}_i - \bf{\tilde{x}}_j  \Vert}^2
\end{eqnarray}
where,
\begin{eqnarray*}
	\bf{\tilde{x}}_i = \left( \rho_1(\textbf{x}_i),...,\rho_k(\textbf{x}_i)  \right) ,  \, \, \, \, \rho_\tau (\textbf{x}_i) = \frac{\displaystyle\sum_{t=1}^{T-\tau} (x_{i_t} - \overline{\textbf{x}}_i) (x_{i_{(t+\tau)}} - \overline{\textbf{x}}_i)}{\displaystyle\sum_{t=1}^T {(x_{i_t} - \overline{\textbf{x}}_i)}^2}
\end{eqnarray*}
and $\tau$ is the time lag.  
So, DACO compares time series by computing the distance between their dynamics, modeled by the auto-correlation operators.
It is worth mentioning that the lower DACO does not represent the similar behavior.

\subsection{Value-based measures}
\label{subsec:valuebased}
In this category,
time series are compared according to their time stamped values. This subsection relies on two standard well-known sub-division: (a) without temporal warping  (e.g. Minkowski distance) and (b)
with considering temporal warping (e.g. Dynamic Time Warping).

\subsubsection{Without temporal warping}
The most used distance function in many applications is the Euclidean (or Euclidian), which is commonly accepted as the simplest distance between sequences. The Euclidean distance $d_E$ ($L_2$ norm) between two time series $\textbf{x}_i=(x_{i1},...,x_{iT})$ and $\textbf{x}_j=(x_{j1},...,x_{jT})$ of length $T$, is defined by:

\begin{eqnarray}
	d_E(\textbf{x}_i,\textbf{x}_j) = \sqrt{\displaystyle\sum_{t=1}^T  {(x_{i_t} - x_{j_t})}^2}
\end{eqnarray}
The generalization of Euclidean Distance is Minkowski Distance, called L$_p$ norm, and  is defined by:

\begin{eqnarray*}
	d_{L_p}(\textbf{x}_i,\textbf{x}_j) = \sqrt[p]{\displaystyle\sum_{t=1}^T  {(x_{i_t} - x_{j_t})}^p}
\end{eqnarray*}
where $p$ is called the Minkowski order. In fact, for Manhattan distance $p = 1$, for the Euclidean distance $p = 2$, while for the Maximum distance $p = \infty$.
All the L$_p$-norm distances do not consider the delay and time warp. Unfortunately, they do not correspond to a common understanding of what a time series (or a sequence) is, and can not capture the flexible similarities.

\subsubsection{With considering temporal warping}
\label{sec:dtw}
Searching the best alignment which matches two time series is a crucial task for many applications. One of the eminent techniques to solve this task is Dynamic Time Warping (DTW); it was introduced in \cite{Bellman,ita75,Kruskall}, with the application in speech recognition. DTW finds the optimal alignment between two time series, and captures the similarities by aligning the elements inside both series. Intuitively, the  series are warped  non-linearly in the time dimension to match each other. Simply, the DTW is a generalization of Euclidean distance which allows a non-linear mapping between two time series by minimizing a global alignment cost between them.
DTW does not require that the two time series data have the same length, and also it can handle the local time shifting by duplicating (or re-sampling) the previous element of the time sequence.

Let  $\textbf{X}=\{\textbf{x}_i\}_{i=1}^N$  be a  set of time series  $\textbf{x}_i=(x_{i1},...,x_{iT})$ assumed of length $T$\footnote{one can make this assumption   as DTW can be applied equally on time series of same or different lengths.}.
 %
An alignment $\normalfont{\bs\pi}$  of length $\lvert \bs\pi  \rvert =m$ between two time series $\normalfont{\textbf{x}_i}$ and $\normalfont{\textbf{x}_j}$ is defined as the set of $m$  ($T \leq m \leq 2 T - 1$) couples of aligned elements of $\normalfont{\textbf{x}_i}$ to elements of $\normalfont{\textbf{x}_j}$:
\begin{center}
$\bs\pi = ( (\pi_1(1), \pi_2(1)), (\pi_1(2), \pi_2(2)),...,(\pi_1(m) , \pi_2(m))) $ 
\end{center}
where $\bs\pi$ defines a warping function that realizes a mapping from time axis of $\normalfont{\textbf{x}_i}$ onto time axis of $\normalfont{\textbf{x}_j}$, and the applications $\pi_1$ and $\pi_2$ defined from $\{1,...,m\}$ to $\{1,..,T\}$ obey to the following  boundary and monotonicity conditions:
\begin{center}
	$1 = \pi_1(1) \leq \pi_1(2) \leq ... \leq \pi_1(m) = T $
	\\ 
	$1 = \pi_2(1) \leq \pi_2(2) \leq... \leq \pi_2(m) = T $		
\end{center}
and  $\forall \, l \in \{1,...,m\}$,   
\bqs
\pi_1(l+1) \leq \pi_1(l) + 1  \; \mbox{and} \; \pi_2(l+1) \leq \pi_2(l) &+& 1, \\
(\pi_1(l+1) - \pi_1(l)) \;  +  \; (\pi_2(l+1) - \pi_2(l)) &\geq& 1
\eqs

\begin{figure}[!ht]
\centerline{
\includegraphics[height= 3.5cm,angle=0]{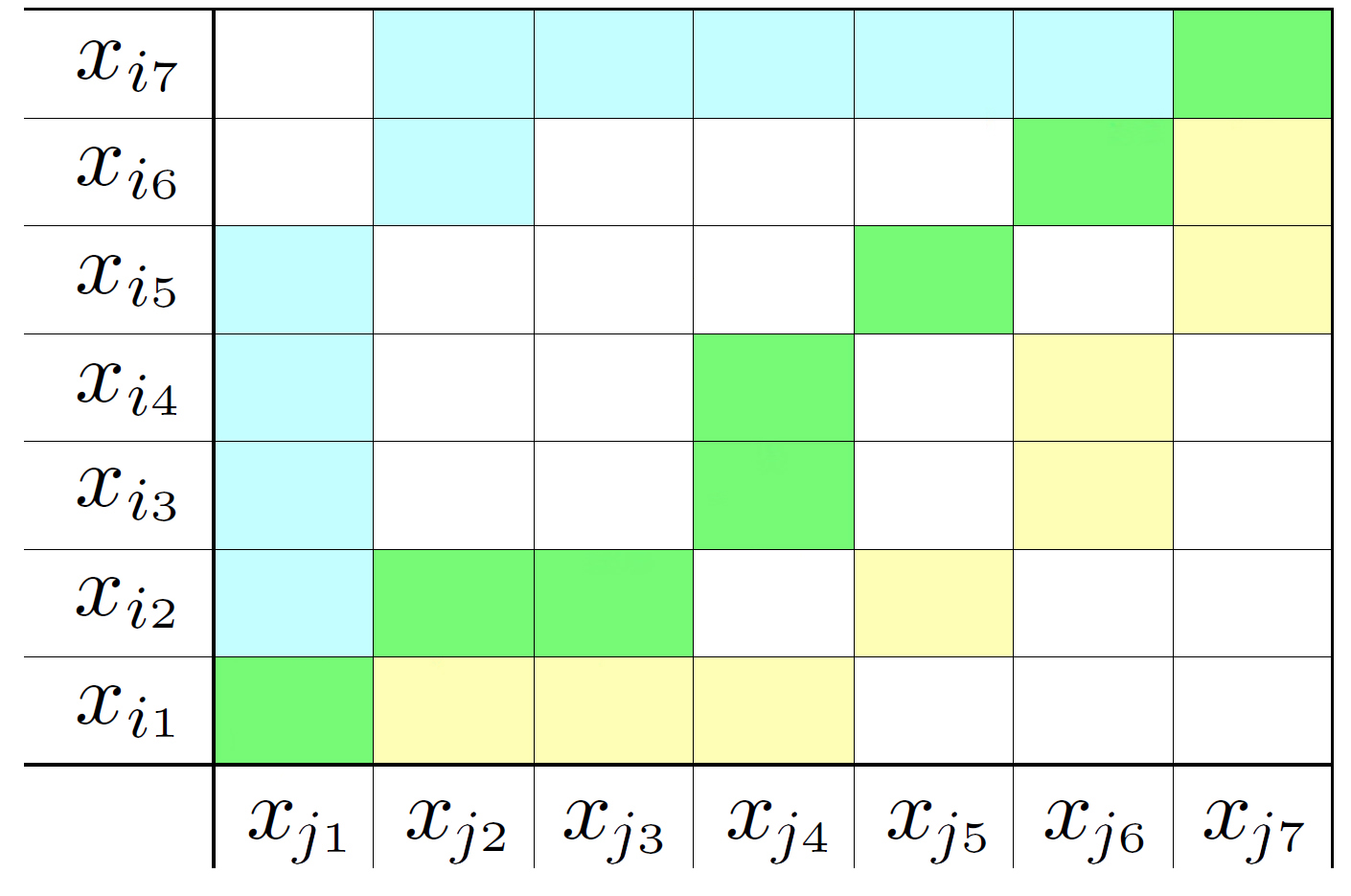}
   }
\caption{\small Three possible alignments path (green, blue, yellow) between  $\textbf{x}_i$ and $\textbf{x}_{j}$}
\label{fig:warppath}
\end{figure}

Intuitively, an alignment $\bs\pi$ defines a way to associate all elements of two time series. The alignments can be described by paths in the  $T \times T$ grid
as displayed in Figure \ref{fig:warppath}, that crosses the elements of time series $\textbf{x}_i$ and  $\textbf{x}_j$. For instance, the green path aligns the two time series as: ((${x}_{i1}$, ${x}_{j1}$), (${x}_{i2}$,${x}_{j2}$), (${x}_{i2}$,${x}_{j3}$), (${x}_{i3}$,${x}_{j4}$), ... , (${x}_{i6}$,${x}_{j6}$), (${x}_{i7}$,${x}_{j7}$)). In the following, we will denote  $\mathcal{A}$ as the set of all possible alignments between two time series.

DTW is currently a well-known dissimilarity measure on time series (and sequences), since it makes them possible to capture temporal distortions.
The DTW between time series $\textbf{x}_i$ and time series $\textbf{x}_j$, with the aim of minimization of the mapping cost, is defined by:
\begin{eqnarray}
\label{definition:dtw}
	\textbf{{\sc dtw}}(\textbf{x}_i,\textbf{x}_j) 	= 	\min_{\bs\pi  \in \mathcal{A}} \,  \sum_{\substack{ (t',t) \in \bs\pi }}   \varphi( x_{it'} , x_{jt} )
\end{eqnarray} 
where $\varphi:  \mathbb{R} \times  \mathbb{R} \rightarrow \mathbb{R^+}$  is a real-valued positive  divergence function (generally Euclidean norm).

While DTW alignments deal with delays or time shifting, the Euclidean alignment $\bs\pi$ between time series $\textbf{x}_i$ and $\textbf{x}_j$ aligns elements observed at the same time:
\begin{eqnarray*}
 \bs\pi = ( (\pi_1(1), \pi_2(1)) , (\pi_1(2), \pi_2(2)) , ... , (\pi_1(T) , \pi_2(T)) )
\end{eqnarray*}
where $\forall t = 1,...,T , \,  \pi_1(t) = \pi_2(t) = t , \, \lvert \bs\pi  \rvert =T$. According to the alignment definition, the euclidean distance ($d_E$) between the time series $\textbf{x}_i$ and $\textbf{x}_j$ is given by:
\begin{eqnarray*}
 d_E(\textbf{x}_i,\textbf{x}_j) \eqdef \displaystyle \displaystyle \sum_{k=1}^{\vert \bs\pi \vert} \varphi( x_{i_{\pi_1(k)}} , x_{j_{\pi_2(k)}} ) = \displaystyle \sum_{t=1}^T \varphi( x_{it} , x_{jt} )
\end{eqnarray*}
where $\varphi$ taken as the Euclidean norm.

\begin{figure}[!ht]
\centering
\captionsetup{justification=centering}
\includegraphics[width=3.2in]{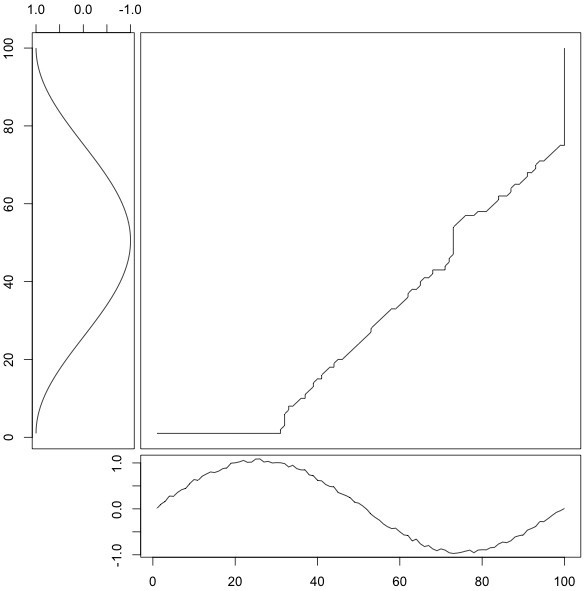} \,
\includegraphics[width=3.2in]{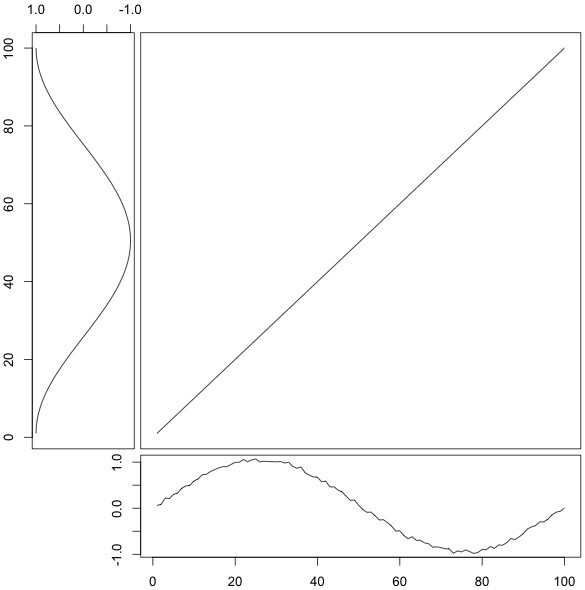}
\caption{\small The optimal alignment path between two sample time series  \\with considering time warp or delay (left), without considering time warp or delay (right)}
\label{fig:edalignment}
\end{figure}

Finally, Figure \ref{fig:edalignment} shows the optimal alignment path between two sample time series with and without considering temporal warping. 
It has not escaped our notice that boundary, monotonicity and continuity are three important properties which are applied to the construction of the path. 
The boundary condition implies that the first elements of the two time series are aligned to each other, as well as the last elements.  
More precisely, the alignment is global and involves the entire time series.
The monotonicity condition preserves the time-ordering of elements. 
In other word, the alignment path does not go back in "\textit{time}" index. 
Lastly, the continuity limits the warping path from long jumps and it guarantees that alignment does not omit important features. 

To find the minimum distance for alignment path, a dynamic programming approach is used. Let  $\textbf{x}_i = (x_{i1},...,x_{it})$ and $\textbf{x}_j = (x_{j1},...,x_{jt'})$  two time series. 
A two-dimensional $\vert \textbf{x}_i \vert$ by $\vert \textbf{x}_j \vert$ cost matrix $D$ is built up, where the value at $D(t,t')$ is the minimun distance warp path that can be constructed from the two time series $\textbf{x}_i$ and $\textbf{x}_j$. Finally, the value at $D(\vert \textbf{x}_i \vert,\vert \textbf{x}_j \vert)$ will contain the minimum distance between the two time series under time warp.  Be aware  that the DTW is not a metric, since it does not satisfy the triangle inequality\footnote{As we can see, for instance, given three time series data, $\textbf{x}_i$ = [0], $\textbf{x}_j$ = [1, 2] and $\textbf{x}_k$ = [2,3,3], then:
 $\textbf{{\sc dtw}}(\textbf{x}_i,\textbf{x}_j) = 3    $ ,
  $\textbf{{\sc dtw}}(\textbf{x}_j,\textbf{x}_k) = 3    $ ,
   $\textbf{{\sc dtw}}(\textbf{x}_i,\textbf{x}_k) = 8    $. Hense,
 $\textbf{{\sc dtw}}(\textbf{x}_i,\textbf{x}_j) + \textbf{{\sc dtw}}(\textbf{x}_j,\textbf{x}_k)  \ngeq  \textbf{{\sc dtw}}(\textbf{x}_i,\textbf{x}_k).$ $\Box$}.

Time and space complexity of DTW are straightforward to derive. Each cell in the cost matrix is filled once in a constant time. The cost of the optimal alignment can be recursively computed by: 
\begin{eqnarray*}
 D(t,t') = d(t,t') + \displaystyle \min \{ D(t-1,t') , D(t-1,t'-1) , D(t,t'-1) \} 
\end{eqnarray*}
where $d$ is a distance function between the elements of time series, and these yield complexity of $O(N^2)$ for the DTW. 
%
Hence, the standard DTW is much too slow for searching the optimal alignment path for large datasets. 
A lot of work has been proposed during the last 50 years to speed-up the DTW measure or variant of it.
In general, the techniques which make DTW faster are divided in three main different categories: 
\begin{enumerate}
\item constraints (e.g. Sakoe-Chiba band \cite{Sakoe71,Sakoe}), 
\item indexing (e.g. piecewise \cite{Keogh2000}),
\item data abstraction (e.g multiscale \cite{Salvador04}). 
\end{enumerate} 

Finally, in spite of the great success of DTW and its variants in a diversity of domains,
there are still several persistent improvement needs about it, such as finding techniques to speed up DTW with no
(or relaxed) constraints and make it more accurate.

\subsubsection{Kernelization of DTW}
Over the last ten years, estimation and learning methods using kernels have become rather popular to cope with non-linearities, particularly in machine learning and data mining. 
More recently, several DTW-based kernels which allow one to process time series with kernel machines have been introduced \cite{Bahlmann,Shimodaira,Cuturi,Marteau2015}, 
and they have been proved useful to handle and analyze structured data such as images, graphs, signals and texts \cite{Hofmann}. 
However, such similarities (e.g. $\mathcal{K}_{\mbox{gdtw}}$, $\mathcal{K}_{\mbox{dtak}}$) cannot be translated easily into positive definite kernels, which is a vital requirement of kernel machines in the training phase \cite{Cuturi,Marteau2015,Joder}. 
Therefore, defining appropriate kernels to handle properly structured objects, and notably
time series, remains a key challenge for researchers.

\cite{Cuturi} introduced the global alignment kernel ($\mathcal{K}_{\mbox{ga}}$), that takes the following form:
\begin{eqnarray}
\label{wkdtw}
	\mathcal{K}_{\mbox{ga}}(\textbf{x}_i,\textbf{x}_j) 	= \sum_{\bs\pi  \in \mathcal{P}} \, \,  \prod_{\substack{ (t,t') \in \bs\pi }} \, \kappa( x_{i_t} , x_{j_{t'}} )
\end{eqnarray}
where $\kappa( . , . )$ is a local kernel and $\mathcal{A}$ the set of all admissible alignment paths.\\

In \cite{Marteau2015} another formal approach has been proposed to derive a positive definite time elastic kernel, $\mathcal{K}_{\mbox{rdtw}}$, close to the DTW matching scheme based on the design of a global alignment positive definite kernel for each single alignment path as given in Eq.\ref{Krdtw}.

\begin{eqnarray}
\label{Krdtw}
	\mathcal{K}_{\mbox{rdtw}}(\textbf{x}_i,\textbf{x}_j) 	= \sum_{\bs\pi  \in \mathcal{P} \subset \mathcal{A}} \, \,  K_\pi(\textbf{x}_i,\textbf{x}_j)
\end{eqnarray}
where this time $\mathcal{P}$  is any subset of the set of all admissible alignment paths $\mathcal{A}$  between two time series, and $K_\pi(\textbf{x}_i,\textbf{x}_j)$ a positive definite kernel associated to path $\pi$ and defined as:

\begin{eqnarray}
\begin{array}{ll}
{\displaystyle K_\pi(\textbf{x}_i,\textbf{x}_j)=\prod_{\substack{(t,t') \in \bs\pi  }} \, \kappa( x_{i_{t}} , x_{j_{t'}} )} \\
\hspace{18mm}+ {\displaystyle \prod_{\substack{(t,t') \in \bs\pi  }} \, \kappa( x_{i_{t'}} , x_{j_{t}} )}\\
\hspace{18mm}+ {\displaystyle \prod_{\substack{(t,t') \in \bs\pi  }} \, \kappa( x_{i_{t}} , x_{j_{t}} )}\\
\hspace{18mm}+ {\displaystyle \prod_{\substack{(t,t') \in \bs\pi  }} \, \kappa( x_{i_{t'}} , x_{j_{t'}} )}\\
\end{array}
\label{kpi}
\label{eq:Kpi}
\end{eqnarray}
with  $\kappa$ a local kernel on $\mathbb{R}^d$ (typically $\kappa(a,b)=e^{-\nu\cdot||a-b||^2}$).\\

Both kernels take into account not only (one of) the best possible alignment, but also all the good (or nearly the best) paths by summing up their global costs. The parameter $\nu$ is used to tune the local matches, thus penalizing more or less alignments moving away from the optimal ones. This parameter can be  optimized by a grid/line search through a cross-validation. \\
 
The main advantage of $\mathcal{K}_{\mbox{rdtw}}$  is that the restriction of the main summation over any subset $\mathcal{P}$ of $\mathcal{A}$ is guaranteed to be positive definite, which is not the case for $\mathcal{K}_{\mbox{ga}}$ (even when $\mathcal{P}$ contains symmetric paths and the main diagonal path). As the sparsification of the DTW path search space involves to restrict the summation on subset of $\mathcal{A}$, this property is quite relevant in an SVM classification context.\\

For both kernels, a dynamic programming computationally efficient solution exists, whose algorithmic complexity is the same as the DTW ($O(T^2)$), although the summation of products of $exp$ functions is more expensive than the evaluation of a $min$ operation.\\

In summary, finding a suitable proximity measure is an important aspect when dealing with
the time series (or sequences),
that captures the essence of the them according to the domain of application. 
For instance, the Euclidean distance is commonly used due to its computational efficiency. 
However, it is very brittle for time series and small shifts of one time series can result in huge distance changes. 
Hence, more sophisticated distances and (dis)similarities have been devised to be more robust to small fluctuations of the input time series. 
Notably, the DTW and its kernalized variant have enjoyed success in many areas where its time complexity is not a critical issue. 

\section{Sparsified-Paths search space  DTW (SP-DTW)}
\label{problem}

The main idea behind this work is to introduce a kind of sparsification of the set of DTW alignment paths to make it faster to evaluate without degrading its accuracy by using an occupancy grid with the weighting values.
The proposition mainly introduces a weighted warping function that constrains the search of the best alignment path in a subset of admissible alignments that has been learned and pruned on the trained data. 

Let $\textbf{X}=\{\textbf{x}_i\}_{i=1}^N$  be a  set of time series  $\textbf{x}_i=(x_{i1},...,x_{iT})$ assumed of length $T$,
 %
and $\normalfont{\bs\pi}_{ij}$  be the optimal alignment between two time series $\normalfont{\textbf{x}_i}$ and $\normalfont{\textbf{x}_j}$ expressed as the set of couples of aligned elements in the $T \times T$ grid.
Let $m_{tt'}^{ij}$ be an indicative variable taking value $1$ if path $\normalfont{\bs\pi}_{ij}$ travels through the grid cell with index $tt'$ ($1 \leq t , t' \leq  T$), $0$ otherwise. Then $p(m_{tt'})$, which represents the normalized frequency of occupancy of $tt'$ cell is defined by:

\begin{equation}
\label{eq:weightSparseMatrix}
p(m_{tt'}) = \frac{\displaystyle\sum_{i,j \in N} m_{tt'}^{ij}}{\displaystyle\sum_{t,t' \in T} \displaystyle\sum_{i,j \in N}  m_{tt'}^{ij}}
\end{equation}

Hence, each cell in the occupancy grid has a value representing the frequency of the occupancy of that cell over all the  optimal alignment paths in the training set. The higher values represent a high likeliness that an  optimal alignment path, randomly drawn from the train set, contains that cell, and
the values close to 0 represent a high likeliness that this optimal alignment path will not cross the cell.
The proposed extended form of DTW, called Sparsified-Paths search space  Dynamic Time Warping (SP-DTW), is then defined as:

\begin{eqnarray}
\label{pdtw}
	{\normalfont {\mbox {SP-DTW}}}(\textbf{x}_i,\textbf{x}_j) 	= 	\min_{\bs\pi  \in \mathcal{P} \subset \mathcal{A}} \, \,  \underbrace{\sum_{\substack{ (t,t') \in \bs\pi }} \,  f(p(m_{tt'})) \, \varphi( x_{i_t} , x_{j_{t'}} )}_{C(\bs\pi)} \\
	= C(\bs\pi^{*}) \nonumber
\end{eqnarray}
where 
$\mathcal{P}$  is any subset of the set of all admissible alignment paths $\mathcal{A}$  between two time series, 
and $\varphi:  \mathbb{R} \times  \mathbb{R} \rightarrow \mathbb{R^+}$ is a positive, real-valued, dissimilarity function (e.g. Euclidean distance).


For function $f(p(m_{tt'}))$, we consider here $p(m_{tt'})^{-\gamma}$, where $\gamma \in \mathbb{R^+} \cup \{0\}$ controls the influence of the weighting scheme.  
The negative exponent guarantees that the most important cells  are privileged in the
optimal alignment that minimizes the SP-DTW. 
For $\gamma = 0$, Eq. \ref{pdtw} leads to the standard DTW.


The cost function $C$ computes the sum of the weighted dissimilarities $\varphi$ between time series  $\normalfont{\textbf{x}_i}$ and $\normalfont{\textbf{x}_j}$ through the alignment $\bs\pi$. 
The fact that $f$ is non-increasing guarantees that the most important cells (i.e. the cells with the higher frequencies) of the grid should be privileged in the optimal alignment that minimizes the cost function $C$. 
Lastly, the optimal alignment path $\bs\pi^*$ is obtained through the same dynamic programming procedure as the one used for the standard DTW.

\begin{figure}[!ht]
\centerline{
\includegraphics[height=9cm,angle=0]{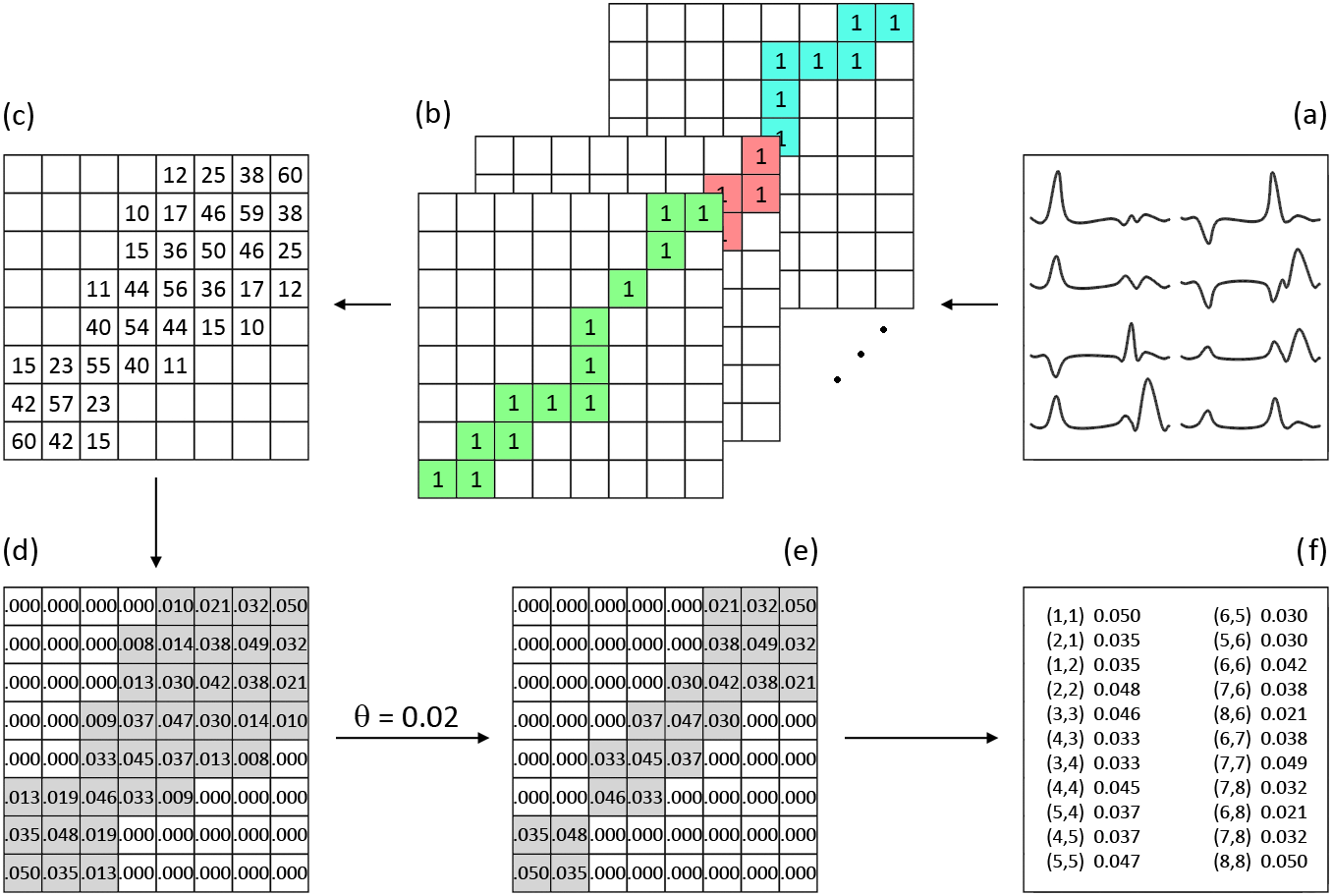}
   }
\caption{\small  The strategy of estimating sparse alignment global  matrix: (a) the time series dataset, (b) the best pairwise alignment paths matrices, (c) the global alignment path matrix (resulting from the summation of all previous matrices), (d) cells that are not visited by an optimal path are set to 0 and the matrix is scaled into [0;1), (e) the cells with an occupancy frequency below $\theta$ are set to 0, (f) the sparse representation of the resulting global alignment matrix.}
\label{fig:PDTW}
\end{figure}
%
%
Figure \ref{fig:PDTW} shows the strategy steps used to estimate the occupancy frequencies. Given a  set of time series (Figure \ref{fig:PDTW}-a), we compute the pairwise alignment path between all the times series in the set. 
The grid cells, which are occupied by the path in each pairwise alignment, get the value 1, and the rest get the value 0 (Figure \ref{fig:PDTW}-b). 
To be more efficient, instead of computing $N^2$ pairwise comparison of DTW on a dataset of $N$ time series, we compute  $N(N-1)/2$ pairs of DTW, then we symmetrize the matrices and  get the same
result as computing the $N^2$ DTW pairwise comparison.
In the next step, we sum all the pairwise Boolean grids to obtain the global absolute frequency symmetric matrix that corresponds to the all of pairwise alignments (Figure \ref{fig:PDTW}-c).   
Then, to get the normalized frequency matrix, we normalize the global matrix in the range of [0 , 1)(Figure \ref{fig:PDTW}-d).
To decrease the time and space complexity, we consider a threshold value $\theta$ for the frequency value of grid cells. The values smaller than the pre-defined threshold will be equal to zero (Figure \ref{fig:PDTW}-e). In practice, $\theta$ is estimated on the train data using a leave one out procedure. Lastly,  we filter out the grid cells with zero weight to obtain a sparsified global matrix (Figure \ref{fig:PDTW}-f).

In practice, the sparsified  alignment path matrix is stored as a list of coordinate (column, row, value) tuples. We denote this list as LOC. The entries of this list are sorted by increasing row index, then increasing column index, such that the sparsified variant of DTW (SP-DTW) can be iteratively evaluated according to Algorithm \ref{dtw-sp}.

\begin{algorithm}[!t]
\caption{SP-DTW}\label{dtw-sp}
\begin{algorithmic}[1]

\Function{\textsc{SP-DTW}}{$X,Y, [W, r_w, c_w]$}
\LeftComment{X,Y: two time series} 
\LeftComment{$[W, r_w, c_w]$ sparse path alignment matrix}
\LeftComment{$W$: weight vector, $r_w$: row index vector, $c_w$: column index vector} 
\State $\textit{L} \gets \text{length of }\textit{W}$
\Comment{$W$, $c_w$ and $r_w$ have the same length}
\State $L_x \gets \text{length of }X$
\State $L_y \gets \text{length of }Y$
\LeftComment{Initialize \textsc{D}, a $(L_x \times L_y)$ matrix, with Max\_Float values}
\State $\textsc{D} \gets Max\_Float\cdot ones(L_x,L_y)$
\State $\textsc{D}(1,1) \gets ||X(1)-Y(1)||^2\cdot W(1)$
\For{$i=2$ to $L$}
\State $ii \gets r_w(i)$
\State $jj \gets c_w(i)$
\If{$jj=1$}
\State $\textsc{D}(ii,1) \gets \textsc{D}(ii-1,1) + ||X(ii)-Y(1)||^2\cdot W(i)$
\ElsIf{$ii=1$}
\State $\textsc{D}(1,jj) \gets \textsc{D}(1,jj-1) + ||X(1)-Y(jj)||^2\cdot W(i)$
\Else
\State $\textsc{D}(ii,jj) \gets ||X(ii)-Y(jj)||^2\cdot W(i) +$ 
\State \hspace{17mm}Min$(\textsc{D}(ii-1,jj-1) ,\textsc{D}(ii-1,jj), \textsc{D}(ii,jj-1))$
\EndIf
\EndFor
\State {\bf{Return}} $\textsc{D}(L_x,L_y)$
\EndFunction
\end{algorithmic}
\end{algorithm}

\section{Sparsified-Paths search space  DTW Kernel (SP-$\mathcal{K}_{\mbox{rdtw}}$)}
\label{kernelproblem}
As already mentioned, we straightforwardly derive a positive definite time elastic kernel for sparsified subsets of alignment paths from $\mathcal{K}_{\mbox{rdtw}}$ itself (Eq. \ref{Krdtw}), where
$\mathcal{P}\subset \mathcal{A}$ in Eq. \ref{Krdtw} determines the sparsity of the path search space. 


\begin{algorithm}[!ht]
\caption{SP-$\mathcal{K}_{\mbox{rdtw}}$}\label{Krdtw-sp}
\begin{algorithmic}[1]
\Function{SP-{\text{$\mathcal{K}_{{rdtw}}$}}}{$X,Y,\nu,r_w,c_w$}
\LeftComment{$X,Y$: two time series}
\LeftComment{$\nu$: $\mathcal{K}_{\mbox{rdtw}}$ parameter}
\LeftComment{We note $\kappa_\nu(x,y)=exp(-\nu\cdot||x-y||^2)$, the local kernel}
\LeftComment{$[r_w, c_w]$ sparse path alignment matrix indices (row and column index vectors)
\State $L \gets \text{length of } r_w}$
\Comment{$c_w$ and $r_w$ have the same length}
\State $L_x \gets \text{length of }X$
\State $L_y \gets \text{length of }Y$
\LeftComment{Initialize $K_1$ and $K_2$, two $(L_x \times L_y)$ matrices, with zero values}
\State $K_1 \gets zeros(L_x,L_y)$
\State $K_2  \gets zeros(L_x,L_y)$
\State $K_1(1,1) \gets \kappa_\nu(X(1),Y(1))$
\State $K_2(1,1) \gets \kappa_\nu(X(1),Y(1))$
\For{$i=1$ to $L$}
\If{($i<L_x$ and $i< L_y$)}
\EndIf
\EndFor
\For{$i=2$ to $L$}
\State $ii \gets r_w(i)$
\State $jj \gets c_w(i)$
\If{$jj=1$}
\State $K_1(ii,1) \gets 1.0/3.0 \cdot K_1(ii-1,1) \cdot \kappa_\nu(X(ii),Y(1))$
\State $K_2(ii,1) \gets 1.0/3.0 \cdot K_2(ii-1,1) \cdot \kappa_\nu(X(ii),Y(ii))$
\ElsIf{$ii=1$}
\State $K_1(1,jj) \gets 1.0/3.0 \cdot K_1(1,jj-1) \cdot \kappa_\nu(X(1),Y(jj))$
\State $K_2(1,jj) \gets 1.0/3.0 \cdot K_2(1,jj-1) \cdot \kappa_\nu(X(jj),Y(jj))$
\Else
\State $K_1(ii,jj) \gets 1.0/3.0 \cdot \kappa_\nu(X(ii),Y(jj)) \cdot$ 
\State \hspace{17mm}$(K_1(ii-1,jj-1) + K_1(ii-1,jj) + K_1(ii,jj-1))$
\State $K_2(ii,jj) \gets 1.0/3.0 \cdot (\kappa_\nu(X(ii),Y(ii))+\kappa_\nu(X(jj),Y(jj)))/2  \cdot K_2(ii-1,jj-1) + $ 
\State \hspace{17mm}$K_2(ii-1,jj)\cdot \kappa_\nu(X(ii),Y(ii)) + K_2(ii,jj-1)\cdot \kappa_\nu(X(jj),Y(jj))$
\EndIf
\EndFor
\State \bf{Return} $K_1(L_x,L_y) + K_2(L_x,L_y)$
\EndFunction
\end{algorithmic}
\label{alg:krdtw}
\end{algorithm}

Note that, as $K_\pi(\textbf{x},\textbf{y})$ is proved to be a positive definite (p.d.) kernel \cite{Marteau2015},  for all subset $\mathcal{P} \subset \mathcal{A}$,  ${\text{SP-}}\mathcal{K}_{\mbox{rdtw}}$, expressed as the sum of p.d. kernels\footnote{\tiny{${\displaystyle K_\pi(\textbf{x},\textbf{y}) = \underbrace{\prod_{\substack{(t,t') \in \bs\pi  }}  \kappa( x_{t} , y_{t'} )
+  \prod_{\substack{(t,t') \in \bs\pi  }}  \kappa( x_{t'} , y_{t} )}_{K_1}
+  \underbrace{\prod_{\substack{(t,t') \in \bs\pi  }}  \kappa( x_{t} , y_{t} )
+  \prod_{\substack{(t,t') \in \bs\pi  }}  \kappa( x_{t'} , y_{t'} )}_{K_2}}$}  \hspace{2mm}(cf. Eq.\ref{eq:Kpi})}, is guaranteed to be p.d., which is not the case for a sparsification of the global alignment kernel proposed in \cite{Cuturi}, even if $\mathcal{P}$ contains symmetric paths (if $\pi \in \mathcal{P}$ then $\overline{\pi} \in \mathcal{P}$, where $\overline{\pi}$ is the symmetric path of $\pi$) and the main diagonal ($(1,1),(2,2) \cdots (T,T) \in \mathcal{P}$).\\

Here again, the sparsified subset of alignment paths $\mathcal{P}$ is characterized by the alignment path matrix that is stored as a list of coordinates (column, row, value) tuples (cf. Fig.\ref{fig:PDTW} (f)). Contrary to the \textsc{SP-DTW} algorithm, for the kernelized version, the weight values are not used, essentially to maintain the definiteness of the sparse kernel. As for SP-DTW the entries of this list are sorted by increasing row index, then increasing column index, such that the sparsified 
$\mathcal{K}_{\mbox{rdtw}}$ kernel can be evaluated iteratively as presented in Algorithm \ref{alg:krdtw}.

While the complexity of computing the DTW is  quadratic in T (the length of the time series),
for Sparse-Paths versions, the complexity is linear with the number of non-zero cells of the path matrix. Hence the algorithmic complexity for \textsc{SP-DTW} and SP-$\mathcal{K}_{\mbox{rdtw}}$ are in between $O(T)$ and $O(T^2)$.
Furthermore, the sparse data is by nature more easily compressed and thus may require significantly less storage. 

\section{Experimental study}
\label{exp}
In this section, we first describe the datasets retained to lead our experiments prior to comparing the Nearest Neighbor (1-NN) and Support Vector Machine (SVM) classification algorithms, based on the proposed (dis)similarity measures 
comparatively with some of the most known and well-used alternative ones.

\subsection{Data sets description}
Our experiments are conducted on 30 public datasets from UCR 
time series classification 
archive \cite{UCRArchive}. 
Table \ref{tab:data} describes the datasets considered, with their main characteristics: number of categories (k), dataset train and test size (N) and time series length (T), where classes composing the datasets are known beforehand.
The datasets retained have very diverse structures and shapes, and as can be seen, they come in  different class numbers, train and test set sizes as well as time series lengths.

\begin{table}[!ht]
\linespread{1}
\setlength{\tabcolsep}{5.5pt}
\centering
\scriptsize
{
\caption{Data description}
\label{tab:data}
\begin{tabular}{@{}lcccc@{}}
\hline
                  & Class Nb. & Size (Train) & Size (Test) &  TS. Length           \\ \cline{2-5}
DataSet           & k         & N            & N           & T        \\   \hline
50Words           & 50        & 450          & 455         & 270  \\
Adiac             & 37        & 390          & 391         & 176 \\
ArrowHead         & 3         & 36           & 175         & 251 \\
Beef              & 5         & 30           & 30          & 470   \\
BeetleFly         & 2         & 20           & 20          & 512   \\
BirdChicken       & 2         & 20           & 20          & 512   \\
Car               & 4         & 60           & 60          & 577 \\
CBF               & 3         & 30           & 900         & 128  \\
ECGFiveDays       & 2         & 23           & 861         & 136     \\
ElectricDevices   & 7         & 8926         & 7711        & 96     \\
FaceFour          & 4         & 24           & 88          & 350   \\
FacesUCR          & 14        & 200          & 2050        & 131   \\
Fish              & 7         & 175          & 175         & 463 \\
FordB             & 2         & 810          & 3636        & 500 \\
Gun-Point         & 2         & 50           & 150         & 150   \\
Ham               & 2         & 109          & 105         & 431   \\
Haptics           & 5         & 155          & 308         & 1092   \\
Herring           & 2         & 64           & 64          & 512   \\
InlineSkate       & 7         & 100          & 550         & 1882   \\
Lighting-2       & 2         & 60           & 61          & 637  \\
Lighting-7       & 7         & 70           & 73          & 319  \\
MedicalImages     & 10        & 381          & 760         & 99     \\
OliveOil          & 4         & 30           & 30          & 570\\
OSULeaf           & 6         & 200          & 242         & 427  \\
ScreenType        & 3         & 375          & 375         & 720  \\
ShapesAll         & 60        & 600          & 600         & 512  \\
SwedishLeaf       & 15         & 500         & 625         & 128  \\
SyntheticControl  & 6         & 300          & 300         & 60  \\
Trace             & 4         & 100          & 100         & 275 \\
Wine              & 2         & 57           & 54          & 234  \\
\hline
\end{tabular}
}
\end{table}

\subsection{Validation protocol}
\begin{figure}[H]
\centering
\captionsetup{justification=centering}
\includegraphics[width=2.33in]{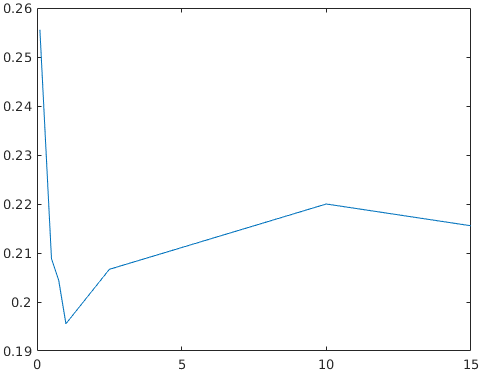} 
\includegraphics[width=2.33in]{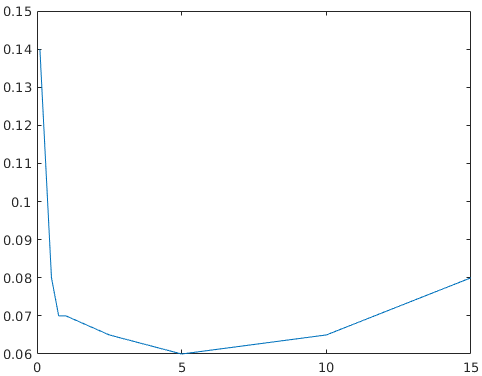} 
\includegraphics[width=2.33in]{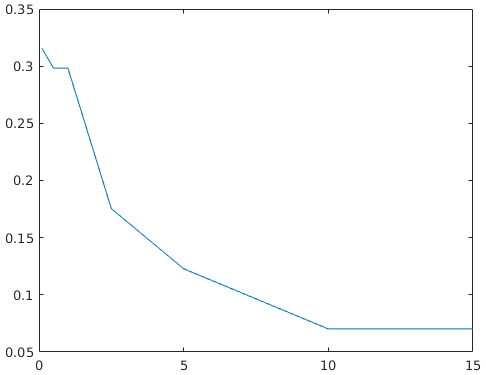} 
\vspace{0.4cm}
\caption{\small The error rate diagram of the leave one out procedure through a grid/line search in the range of  $[0,15]$ on the train dataset 50words (left), facesUCR (middle), and wine (right) 
}
\label{fig:thresholderrors}
\end{figure}

Here we compare the 1-NN and SVM classification algorithms based on the proposed (dis)similarity measures (SP-DTW and SP-$\mathcal{K}_{\mbox{rdtw}}$) with the classical CORR, DACO,  Ed, DTW  measures as well as the DTW$_{sc}$ and $\mathcal{K}_{\mbox{rdtw}}$.
We focus on the above measures because they constitute the most frequently
used proximity measure and distance metrics in temporal data mining. 
For our comparison in this paper, we rely on the classification error rate, which measures the agreement between the predicted class labels and the actual ones, to evaluate each metric, and the time speed-up percentage. 
The classification error rate lies in [0, 1] and the lower index, the better the agreement is. 
In particular, the best value of error rate, 0, is reached when the predicted class labels are equivalent to the actual class labels. 
Finally, the meta parameters used in some of the measures (e.g. $\tau$,  the time lag in DACO, $\theta$, the threshold value in SP-DTW, or $\nu$ in $\mathcal{K}_{\mbox{rdtw}}$) are selected according to  a k-fold cross validation set or a leave one out procedure through a standard grid search process carried out on the train data.
For instance, Figure \ref{fig:thresholderrors} shows the threshold estimation for some sample dataset using the error rate of leave one out procedure through a grid/line search. The lower error rate value indicates the optimal threshold.

\subsection{Experimental results}
The error rate for 1-NN classification for each distance or (dis)similarity, on each dataset, are reported in Table \ref{tab:1-NN}. Results in bold correspond to
the best values.

\begin{table}[!ht]
\linespread{1}
\setlength{\tabcolsep}{5.8pt}
\centering
\scriptsize
{
\caption{Comparison of 1-NN classification error rate}
\label{tab:1-NN}
\begin{tabular}{@{}lcc|cccc||cc@{}}
\hline
                  &  \multicolumn{7}{c}{1-NN (Nearest Neighbor)}            \\ \cline{2-9}
                  &  \multicolumn{2}{c|}{\sc behavior-based}   &  \multicolumn{6}{c}{\sc value-based}         \\ \cline{2-3} \cline{4-9}
DataSet           &  CORR    &   DACO    &   Ed  & DTW   & DTW$_{sc}$ & Krdtw  & SP-DTW  & SP-Krdtw  \\   \hline
50Words           &  0.369   &  0.261    & 0.369 &  0.310 &  0.242(6) & \textbf{0.196}  &   0.226    & 0.202 \\ 
Adiac             &  0.389   &  0.445    & 0.389 &  0.396 & 0.391(3)   & 0.379  & 0.391   &  \textbf{0.368}   \\
ArrowHead         &  0.200  &  0.234    & 0.200 & 0.297 & 0.200(0)  &  0.200  &  0.217  & \textbf{0.194} \\
Beef              &  0.333   &  0.367    & 0.333 & 0.367 & 0.333(0)   & 0.367   & \textbf{0.267}   &  0.333  \\
BeetleFly         &  \textbf{0.250}   &  \textbf{0.250}    & \textbf{0.250} & 0.300 &  0.300(7) & 0.300  &    \textbf{0.250}   & 0.300  \\
BirdChicken       &  0.450   &  0.300    & 0.450 & 0.250 &  0.300(6)   & 0.250  & 0.250  & \textbf{0.150} \\
Car               &  0.267   &  0.283    & 0.267 & 0.267 &  0.233(1) &  \textbf{0.183} &   \textbf{0.183}  &  \textbf{0.183} \\
CBF               &  0.148   &  0.232    & 0.148 & 0.003 &   0.004(11)   & \textbf{0.002}  & 0.014  & \textbf{0.002} \\
ECGFiveDays       &  0.203   &  0.202    & 0.203 & 0.232 & 0.203(0)  & 0.124  &   0.189  &  \textbf{0.091}  \\
ElectricDevices   &  0.450   &  0.434    & 0.450 & 0.399 &   \textbf{0.376}(14) & 0.384  &  0.394 &  \textbf{0.376}   \\
FaceFour          &  0.216   &  0.193    & 0.216 & 0.170 & 0.114(2)  & 0.114 &   0.114   & \textbf{0.102} \\
FacesUCR          &  0.231   &  0.192    & 0.231 & 0.095 &   0.088(12) & \textbf{0.067} &   0.080 &  \textbf{0.067}   \\
Fish              &  0.217   &  0.265    & 0.217 & 0.177 & 0.154(4)  & \textbf{0.086}   &   0.177 &  \textbf{0.086}  \\
FordB             &  0.442   &  \textbf{0.261}    & 0.442 & 0.406 & 0.414(1)  & 0.410  &   0.395  &  0.413  \\
Gun-Point         &  0.087   &  0.067    & 0.087 & 0.093 &   0.087(0) & \textbf{0.027}  &   \textbf{0.027} & \textbf{0.027}  \\
Ham               & 0.400  &  0.514    & 0.400 & 0.533 & 0.400(0)     & 0.400  &  0.400 & \textbf{0.371} \\
Haptics           &  0.630   &  0.636    & 0.630 & 0.623 &  0.588(2)   &  \textbf{0.471} & 0.584  & 0.542 \\
Herring           &  0.484   &  0.500    & 0.484 & 0.469 & 0.469(5)    & 0.406 &  0.421  & \textbf{0.375} \\
InlineSkate       &  0.658   &  0.680    & 0.658 & 0.616 & 0.613(14)  & \textbf{0.562}  &  0.592  &  \textbf{0.562}  \\
Lighting-2       &  0.246   &  0.197    & 0.246 & 0.131 &  0.131(6)   & \textbf{0.098}  & 0.114 &  \textbf{0.098}  \\
Lighting-7       &  0.425   &  0.382    & 0.425 & 0.274 &   0.288(5)  & 0.247  &  0.260 & \textbf{0.219}   \\
MedicalImages     &  0.316   &  0.330    & 0.316 & 0.263 & 0.253(20)   & 0.265  & 0.257  & \textbf{0.240}  \\
OliveOil          &  \textbf{0.133}   &  0.167    & \textbf{0.133} & 0.167 & \textbf{0.133}(0)  & \textbf{0.133}  &   \textbf{0.133}   & 0.167 \\
OSULeaf           &  0.479   &  0.422    & 0.479 & 0.409 &  0.388(7)  & 0.306  &  0.384   & \textbf{0.302} \\
ScreenType        &  0.640   &  0.669    & 0.640 & 0.603 &  \textbf{0.589}(17) &  0.613  &  \textbf{0.589} &  0.592 \\
ShapesAll         &  0.248   &  0.280    & 0.248 & 0.232 &  0.198(4)  & 0.177 &   0.223  & \textbf{0.173} \\
SwedishLeaf       &  0.211   &  0.162    & 0.211 & 0.208 &  0.154(2)  & 0.158  &   \textbf{0.123}  & 0.137  \\
SyntheticControl  &  0.120   &  0.353    & 0.120 & 0.007 &   0.017(6) &  \textbf{0.003}  &   0.010  &   0.007 \\
Trace             &  0.240   &  0.140    & 0.240 & \textbf{0.000} &  0.010(3)  & 0.010  &   \textbf{0.000}  & 0.010  \\
Wine              &  0.389   &  0.407    & 0.389 & 0.426 & 0.389(0)  & 0.370  &  \textbf{0.351}  &  \textbf{0.351}   \\
\hline
Mean rank & 5.37 & 6.27  & 5.37 & 5.03 & 3.53 & 2.30 & 2.63 & \textbf{1.73}\\
\hline
\end{tabular}
}
\end{table}

Based on the classification error rate displayed in Table \ref{tab:1-NN}, one can note that 1-NN classification with SP-$\mathcal{K}_{\mbox{rdtw}}$ leads to the best classification results overall (20 datasets out of 30). This is highlighted in the last row of the table which presents the mean rank position for each tested method. 
In particular, one shown the DTW with Sakoe-Chiba corridor is outperformed by our two proposed measures.
Furthermore, from Table \ref{tab:1-NN}, one can experimentally verify that the 1-NN classification based on the Pearson correlation coefficient (CORR) has the same error rate with the 1-NN based on the Euclidean distance (Ed). This is theoretically verified in the Appendix A.

\begin{table}[!ht]
\setlength{\tabcolsep}{10pt}
\centering
\scriptsize
{
\caption{ Wilcoxon signed-rank test of pairwise error rate differences for 1-NN}
{\begin{tabular}{|c|c|c|c|c|c|c|}
\hline
Method    &	DACO	&	DTW	&  DTW$_{sc}$	&	$\mathcal{K}_{\mbox{rdtw}}$	&	SP-DTW	&	SP-$\mathcal{K}_{\mbox{rdtw}}$  \\ \hline
\hline 
CORR/Ed & 0.9225 &  0.0125 & \textbf{p$<$0.0001} & \textbf{p$<$0.0001} & \textbf{p$<$0.0001}  &  \textbf{p$<$0.0001}   \\ \hline
DACO & - & 0.0296 & \textbf{0.0002} & \textbf{p$<$0.0001} & \textbf{p$<$0.0001} &  \textbf{p$<$0.0001}\\ \hline
DTW  & - & - & \textbf{0.0012} & \textbf{p$<$0.0001} & \textbf{p$<$0.0001} &  \textbf{p$<$0.0001}    \\ \hline 
DTW$_{sc}$	 & - & - & - & \textbf{0.0012} & \textbf{0.0068}   &  \textbf{0.0001} \\ \hline
$\mathcal{K}_{\mbox{rdtw}}$  & - & - & - & - & \textbf{0.0011} &   0.0189  \\ \hline
SP-DTW  & - & - & - & - &  -  &  0.0249 \\ \hline
\end{tabular}}
\label{tab:significance_1NN}
}
\end{table}

In Table \ref{tab:significance_1NN} we report the P-values for each pair of tested algorithms using a Wilcoxon  signed-rank test. The null hypothesis is that for a tested pair of classifiers, the difference between classification error rates obtained on the 30 datasets follows a symmetric distribution around zero. With a $0.01$ significance level, the P-values that lead to reject the null hypothesis are shown in bolded fonts in the table. It turns out that SP-$\mathcal{K}_{\mbox{rdtw}}$ performs significantly better than all the other tested measures except for $\mathcal{K}_{\mbox{rdtw}}$ since  the differences between them is not significant according to the Wilcoxon signed-rank test. Furthermore, the differences between  SP-$\mathcal{K}_{\mbox{rdtw}}$ and SP-DTW are not significant, but both measures are significantly better than DTW$_{sc}$ and the remaining measures (except $\mathcal{K}_{\mbox{rdtw}}$). Hence, the learned sparsification of alignment paths outperforms significantly the well-used Sakoe-Schiba DTW with adjusted corridor.

\begin{table}[!ht]
\linespread{1}
\setlength{\tabcolsep}{10pt}
\centering
\scriptsize
{
\caption{Comparison of SVM classification error rate}
\label{tab:svm}
\begin{tabular}{@{}lccc||c@{}}
\hline
                  &  \multicolumn{4}{c}{SVM (Support Vector Machine)}            \\ \cline{2-5}
DataSet           &  Ed    &  $\mathcal{K}_{\mbox{rdtw}}$  & $\mathcal{K}_{{\mbox{rdtw}}_{sc}}$ &  SP-$\mathcal{K}_{\mbox{rdtw}}$  \\   \hline
50Words           &  0.301   & \textbf{0.182}   &  0.215    & 0.213           \\ 
Adiac             &  0.258   & \textbf{0.238}   &  0.281    & 0.260           \\ 
ArrowHead         &  \textbf{0.160}   & 0.183   &  0.183    & 0.166           \\ 
Beef              &  \textbf{0.167}   & 0.233   &  0.233    & 0.200           \\ 
BeetleFly         &  0.250   & \textbf{0.100}   &  0.150    & 0.300           \\ 
BirdChicken       &  0.300   & \textbf{0.200}   &  0.300    & 0.350           \\ 
Car               &  0.200   & 0.217   &  0.200    & \textbf{0.133}           \\ 
CBF               &  0.116   & \textbf{0.002}   &  0.004    & \textbf{0.002}           \\ 
ECGFiveDays       &  0.028   & \textbf{0.019}   &  0.072    & 0.022           \\ 
ElectricDevices   &  0.406   & \textbf{0.280}   &  0.365    & 0.365           \\ 
FaceFour          &  0.182   & 0.102   &  0.125    & \textbf{0.091}           \\ 
FacesUCR          &  0.195   & 0.049   &  \textbf{0.045}    & 0.053           \\ 
Fish              &  0.126   & \textbf{0.046}   &  0.063    & 0.053           \\ 
FordB             &  0.392   & \textbf{0.139}   &  0.247    & 0.147           \\ 
Gun-Point         &  0.053   & \textbf{0.007}   &  \textbf{0.007}    & \textbf{0.007}           \\ 
Ham               &  0.295   & 0.324   &  0.476    & \textbf{0.229}           \\ 
Haptics           &  0.552   & \textbf{0.435}   &  0.532    & 0.448           \\ 
Herring           &  0.406   & 0.422   &  0.375    & \textbf{0.344}           \\ 
InlineSkate       &  0.731   & 0.600   & \textbf{0.576}    & 0.684           \\ 
Lighting-2        &  0.279   & \textbf{0.148}  &  0.180    & \textbf{0.148}           \\ 
Lighting-7        &  0.384   & \textbf{0.178}   &  0.247    & \textbf{0.178}           \\ 
MedicalImages     &  0.287   & \textbf{0.229}   &  0.286    & 0.276           \\ 
OliveOil          &  0.133   & 0.133   &  0.133    & \textbf{0.100}           \\ 
OSULeaf           &  0.438   & \textbf{0.236}   &  0.293    & 0.252           \\ 
ScreenType        &  0.619   & 0.600   &  0.656    & \textbf{0.597}           \\ 
ShapesAll         &  0.218   & \textbf{0.160}   &  0.185    & 0.172           \\ 
SwedishLeaf       &  0.096   & 0.054   &  \textbf{0.053}    & 0.109           \\ 
SyntheticControl  &  \textbf{0.020}   & 0.033   &  0.040    & \textbf{0.020}           \\ 
Trace             &  0.240   & \textbf{0.000}   &  0.053    & \textbf{0.000}           \\ 
Wine              &  0.148   & 0.333   &  0.407    & \textbf{0.111}           \\ 
\hline
Mean rank & 3.13 & \textbf{1.73} & 2.73 & 1.87\\
\hline
\end{tabular}
}
\end{table}

In addition, Table \ref{tab:svm} displays the performance of SVM classification over each measure on each dataset for further experiments. Similar as before, results in bold correspond to the best values.
As can be seen, based on the SVM classification error rate, SVM with $\mathcal{K}_{\mbox{rdtw}}$ leads to the best classification results overall (17 datasets out of 30), followed by SVM classification with SP-$\mathcal{K}_{\mbox{rdtw}}$ (13 datasets out of 30). Here again, this is highlighted in the last row of the table which presents the mean rank position for each tested method. One can see that the 
$\mathcal{K}_{{\mbox{rdtw}}_{sc}}$
(that involves a fixed-size Sakoe-Chiba corridor) is outperformed by our proposed measure SP-$\mathcal{K}_{\mbox{rdtw}}$. 
To verify for significant differences, as above, we used the Wilcoxon signed-rank tests.

\begin{table}[!ht]
\setlength{\tabcolsep}{14.3pt}
\centering
\scriptsize
{
\caption{Wilcoxon signed-rank test of pairwise error rate differences for SVM}
{\begin{tabular}{|c|c|c|c|}
\hline
Method           & $\mathcal{K}_{\mbox{rdtw}}$ & $\mathcal{K}_{{\mbox{rdtw}}_{sc}}$ & SP-$\mathcal{K}_{\mbox{rdtw}}$  \\ \hline
\hline 
 Ed     & \textbf{p$<$0.0001} & {0.0279} & \textbf{0.0002} \\ \hline
$\mathcal{K}_{\mbox{rdtw}}$    & -       & \textbf{0.0002} & 0.5539 \\ \hline
$\mathcal{K}_{{\mbox{rdtw}}_{sc}}$    &  -      & -  & \textbf{0.0097}  \\ \hline 
\end{tabular}}
\label{tab:significance_SVM}
}
\end{table}

Table \ref{tab:significance_SVM} gives the P-values for the Wilcoxon  signed-rank tests.  With the same null hypothesis as above (difference between the error rates follows a symmetric distribution around zero), and with a $0.01$ significance level, the P-values that lead to reject the null hypothesis are presented in bolded fonts in the tables. These tests show that no significance difference exists between $\mathcal{K}_{\mbox{rdtw}}$ and SP-$\mathcal{K}_{\mbox{rdtw}}$, while $\mathcal{K}_{\mbox{rdtw}}$ and SP-$\mathcal{K}_{\mbox{rdtw}}$ are significantly better than Ed and $\mathcal{K}_{{\mbox{rdtw}}_{sc}}$. Here again, the learned sparsification of alignment paths significantly outperforms the Sakoe-Chiba variant of the kernelized DTW ($\mathcal{K}_{{\mbox{rdtw}}_{sc}}$) with adjusted corridor on the train data.

As far as one can see in Table \ref{tab:1-NN} until \ref{tab:significance_SVM}, the best classification results are obtained by SP-$\mathcal{K}_{\mbox{rdtw}}$, SP-DTW and  $\mathcal{K}_{\mbox{rdtw}}$.
The main benefit is an important speed-up without loss of accuracy. 
To justify our speed-up claim, Table \ref{tab:complexity} shows the complexity, expressed in terms of number of cell visited in the alignment path matrix, of the proposed paths search space sparsified measures in comparison with the standard DTW and its kernelization.

To do so, we consider the number of visited cells in the alignment path matrix, in each pairwise comparison, for each measure. While for standard DTW (or its kernelization), we need to consider the whole grid cells ($T^2$) for each time series pairwise comparison, the results illustrate that we dramatically decreased the number of visited cells for each pairwise comparison in SP-DTW and SP-$\mathcal{K}_{\mbox{rdtw}}$.
To justify, for each measure, we obtain the speed-up percentage (S), which is the number of visited cells divided by the total number of grid cells ($T^2$). 
As can be seen, for the proposed measure, we decline the cells visited numbers more than 82$\%$ in average, while the speed-up is almost 89\% for DTW$_{sc}$ or $\mathcal{K}_{\mbox{rdtw}_{sc}}$.



\begin{table}[H]
\linespread{1}
\setlength{\tabcolsep}{3.5pt}
\centering
\scriptsize
{
\caption{Comparison of time speed-up (in $\%$) comparatively to the standard DTW (or $\mathcal{K}_{\mbox{rdtw}}$)}
\label{tab:complexity}
\begin{tabular}{@{}lc|cc|cc|cc@{}}
\hline   
                  & DTW/$\mathcal{K}_{\mbox{rdtw}}$    &\multicolumn{2}{c|}{DTW$_{sc}$/$\mathcal{K}_{\mbox{rdtw}_{sc}}$}   & \multicolumn{2}{c|}{SP-DTW}  & \multicolumn{2}{c}{SP-$\mathcal{K}_{\mbox{rdtw}}$}           \\ \cline{2-8}
DataSet           & $\#$ visited cells       & $\#$ visited cells   & S($\%$)         & $\#$ visited cells    & S($\%$)        & $\#$ visited cells  & S($\%$)       \\   \hline
50Words           & 72,900        &  \textbf{8,638}    & \textbf{88.1}  &   12,798   & 82.4    &  12,682  & 82.6   \\
Adiac             & 30,976        &  1,906  & 93.8    &   \textbf{1,320}   & \textbf{95.7}    &   1,324   & \textbf{95.7}   \\
ArrowHead         &  63,001       &   \textbf{251}  & \textbf{99.6}   &   4,233     & 93.3    &   3,793  & 94.0    \\
Beef              & 220,900        &  \textbf{470}   &  \textbf{99.8}   &  11,974    & 94.6    &  10,973  & 95.0   \\
BeetleFly         & 262,144        & 35,092    & 86.6   &    15,760  & 93.9    &  \textbf{14,783}   & \textbf{94.4}   \\
BirdChicken       &  262,144       &  \textbf{30,302}   & \textbf{88.4}    &     61,738  & 76.4   &  72,438    & 72.4   \\
Car               & 332,929        &  \textbf{6,317}   & \textbf{98.1}    &  34,851     & 89.2   &  34,283   & 89.7    \\
CBF               &  16,384       &  \textbf{3,502}   & \textbf{78.6}    &  7,188    & 56.1    &    7,090   & 56.7  \\
ECGFiveDays       &  18,496       &   \textbf{136}   &  \textbf{99.2}    & 2,288     & 87.6    &    2,170  & 88.3   \\
ElectricDevices   & 9,216        &   2,410   & 73.8  &    \textbf{1,224}    & \textbf{86.7}   &  1,228   & \textbf{86.7}    \\
FaceFour          &  122,500       &   \textbf{5,194}   & \textbf{95.7}    &  7,162    & 94.2    &  6,539  & 94.7     \\
FacesUCR          &  17,161       &  3,821    & 77.7  &   3,413    & 80.1   &    \textbf{3,197}   & \textbf{81.4}    \\
Fish              &  214,369       &  \textbf{16,789}  & \textbf{92.2}     & 17,729   & 92.0     &  18,531  & 91.4    \\
FordB             &  250,000       &  \textbf{5,470}  & \textbf{97.8}    &   55,722   & 77.7    &   55,254    & 77.9  \\
Gun-Point         &  22,500       &   \textbf{150}  & \textbf{99.3}    & 5,140   & 77.2      &    5,159   & 77.1  \\
Ham               &  185,761       &   \textbf{431}  & \textbf{99.7}    &    10,581   & 94.3   &   10,375 & 94.4     \\
Haptics           &  1,192,464       &   \textbf{46,494}   & \textbf{96.1}   &   263,322   & 77.9    &  260,049  & 78.2     \\
Herring           & 262,144        &  \textbf{25,462}   & \textbf{90.3}   &  27,256   & 89.6     &   27,002  & 89.7    \\
InlineSkate       &  3,541,924       &  922,382    & 73.9  &   \textbf{560,484}  &  \textbf{84.2}     &  689,409   & 80.5    \\
Lighting-2       &  405,769       &  \textbf{47,567}    & \textbf{88.3}   & 78,789   &   80.6   &  89,547    &  77.9 \\
Lighting-7       &  101,761       &  \textbf{9,649}   & \textbf{90.5}    & 26,435   & 74.0      &   27,019  & 73.4    \\
MedicalImages     &  9,801       &   \textbf{3,481}  & \textbf{64.4}    &    5,101  & 48.0    &   5,085   & 48.1   \\
OliveOil          &  324,900       &   \textbf{570}   & \textbf{99.8}   &    3,520    & 98.9  &   3,747 & 98.8     \\
OSULeaf           &  182,329       &   \textbf{24,323}   & \textbf{86.7}   & 61,045     & 66.5    &  61,757   & 66.1    \\
ScreenType        &  518,400       &  \textbf{161,394}  & \textbf{68.9}     &     274,048  & 47.1   &  279,345   & 46.1    \\
ShapesAll         &  262,144       &  \textbf{20,572}   & \textbf{92.1}    & 33,146     & 87.4    & 30,185   & 88.5   \\
SwedishLeaf       &  16,384      &  \textbf{634}  & \textbf{96.1}     &  1,100   & 93.3     &    1,096  & 93.3     \\
SyntheticControl  &  3,600       &  \textbf{408}   & \textbf{88.7}    &  610    & 83.1    &  674   & 81.3   \\
Trace             &   75,625      &  \textbf{4,603}   & \textbf{93.9}    &  17,263    & 77.2    &    17,529   & 76.8  \\
Wine              &  54,756       &   \textbf{234}  & \textbf{99.6}    &     992    & 98.2 &    1,072   & 98.0  \\ \hline
\multicolumn{2}{l}{\hspace{-0.06cm}Average (speed-up percentage)} & & \textbf{89.9} &  & 82.6 & & 82.3 \\
\hline
\end{tabular}
}
\end{table}

\begin{figure}[H]
\centering
\captionsetup{justification=centering}
\includegraphics[width=2in]{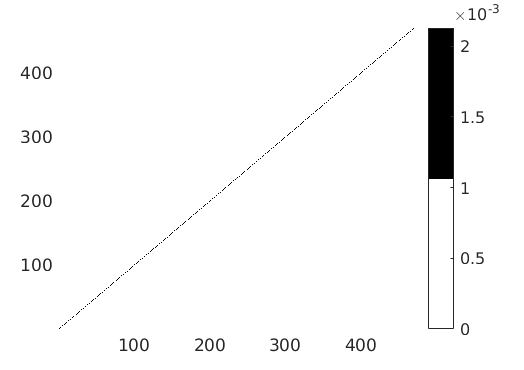} 
\includegraphics[width=2in]{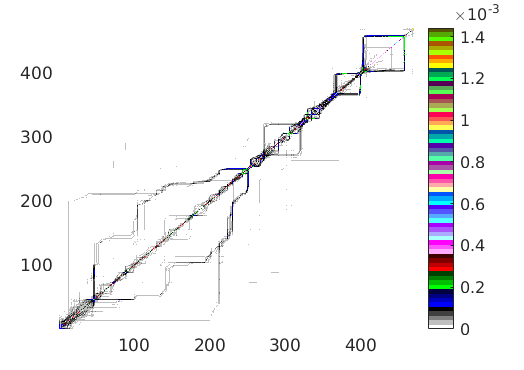} 
\includegraphics[width=2in]{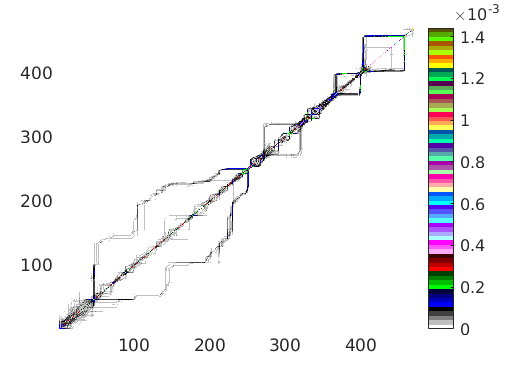} 
\vspace{-0.2cm}
\caption{\small Beef: color-coded of grid cells DTW with optimal Sakoe-Chiba band (left), with sparse paths DTW (middle), with  sparse paths DTW considering a threshold (right)}
\label{fig:comp_cells_beef}
\end{figure}

\begin{figure}[H]
\centering
\captionsetup{justification=centering}
\includegraphics[width=2in]{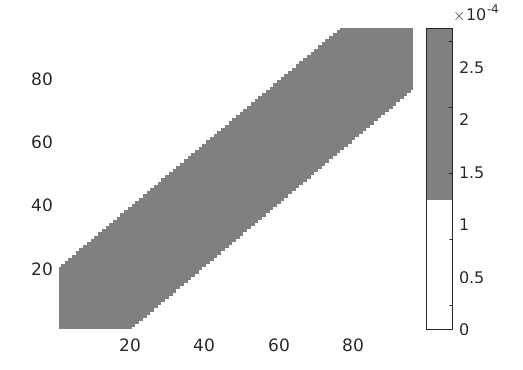} 
\includegraphics[width=2in]{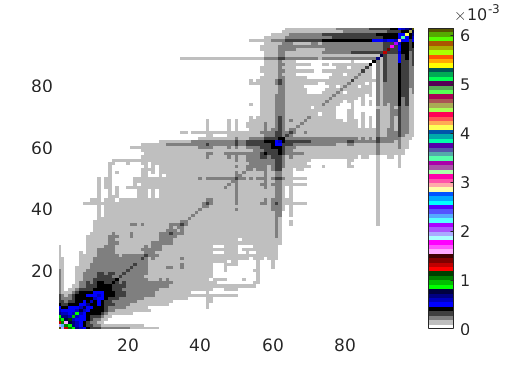} 
\includegraphics[width=2in]{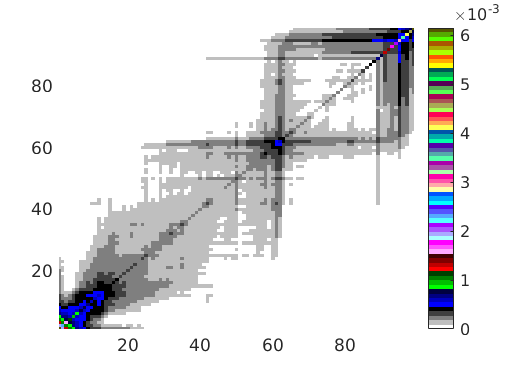}  
\vspace{-0.2cm}
\caption{\small MedicalImages:  color-coded of grid cells DTW with optimal Sakoe-Chiba band (left), with sparse paths DTW (middle), with  sparse paths DTW considering a threshold (right)}
\label{fig:comp_cells_medicalimages}
\end{figure}

Lastly, to have a closer look at the grid cells search space in time series pairwise comparison, in Figures \ref{fig:comp_cells_beef} till \ref{fig:comp_cells_medicalimages}, we plot the color-coded of occupancy grid cells of some sample datasets for the  DTW (or $\mathcal{K}_{\mbox{rdtw}}$) with the optimal Sakoe-Chiba band and for the path sparsified measures with and without considering the threshold value,  according to the frequency of occupancy over the optimal alignment paths in the training set.

\begin{figure}[H]
\centering
\captionsetup{justification=centering}
\includegraphics[width=2in]{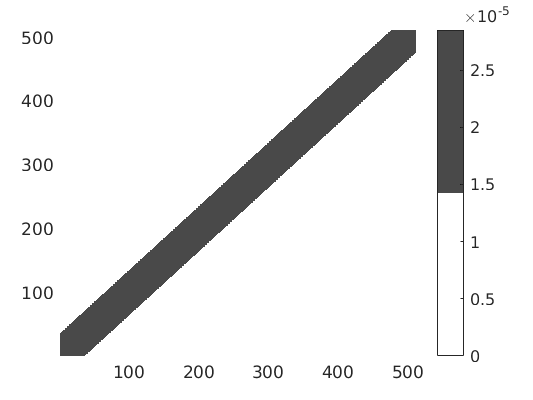} 
\includegraphics[width=2in]{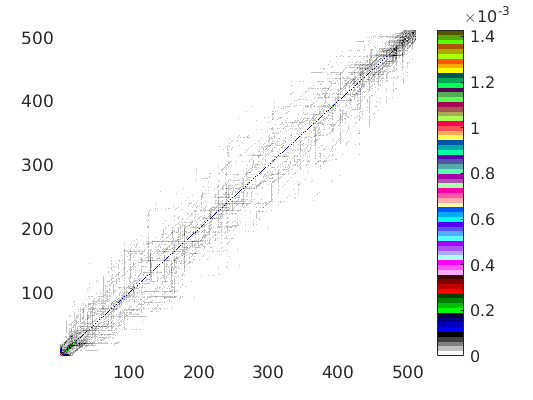} 
\includegraphics[width=2in]{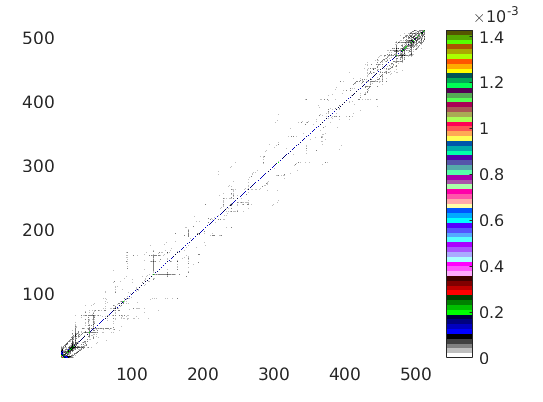} 
\vspace{-0.2cm}
\caption{\small BeetleFly:  color-coded of grid cells DTW with optimal Sakoe-Chiba band (left), with sparse paths DTW (middle), with  sparse paths DTW considering a threshold (right)}
\label{fig:comp_cells_beetlefly}
\end{figure}

\begin{figure}[H]
\centering
\captionsetup{justification=centering}
\includegraphics[width=2in]{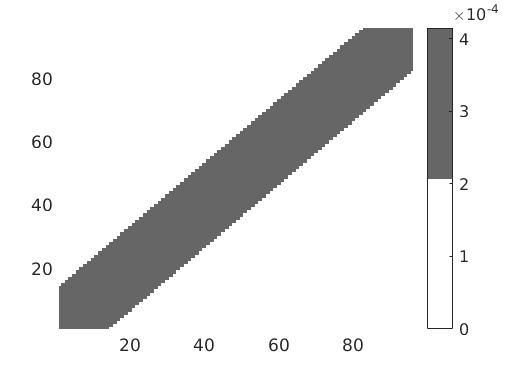} 
\includegraphics[width=2in]{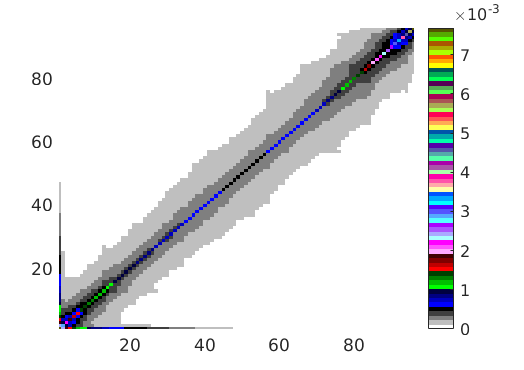} 
\includegraphics[width=2in]{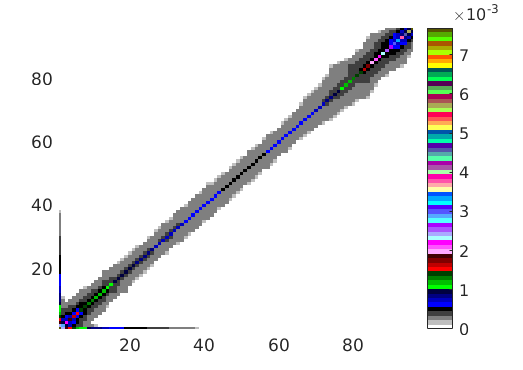} 
\vspace{-0.2cm}
\caption{\small ElectricDevices:  color-coded of grid cells DTW with optimal Sakoe-Chiba band (left), with sparse paths DTW (middle), with  sparse paths DTW considering a threshold (right)}
\label{fig:comp_cells_electricdevices}
\end{figure}

\section{Conclusion}
\label{con}
The Dynamic Time Warping (DTW) is among the most commonly used measure for temporal data comparison in several domains such as signal processing, image processing and pattern recognition.
This research work proposes a sparsification of the alignment path search space for DTW-like measures and their kernelization $\mathcal{K}_{\mbox{rdtw}}$. This led us to propose two new (dis)similarity measures for temporal data analysis and comparison under time warp that cope with such kind of sparsity.
For this, we propose an extension of the standard DTW and its kernelization, based on a weighting, learned over the optimal alignment paths in the training set and  the sparsification of the admissible alignment path search space using a threshold on these weights. 
The efficiency of the proposed (dis)similarity measures, has been analyzed on a wide range of public datasets on two classification tasks: 1-NN and SVM (for the definite measures).
The results show that the classification with path sparsified measures leads to an important speed-up without losing on the accuracy. In particular, these measures significantly outperform  the DTW with a Sakoe-Chiba corridor whose size is optimized on the train data. Finally, we have shown that the kernelized version of DTW admits a straightforward path sparsification variant, much faster and with similar accuracy. 

%

\newpage

\section*{\sc Appendix A}
\label{appendixB}
The Pearson correlation coefficient between time series \textbf{x} and \textbf{y} are defined as follows:

\begin{equation}
\label{appcorr}
{\small
\textbf{{\sc corr}}(\textbf{x},\textbf{y}) = \frac{\textbf{{\sc cov}}(\textbf{x},\textbf{y})}{\sigma(\textbf{x}) \sigma(\textbf{y})}
}
\end{equation}

Since, the numerator of the equation (COV), covariance of two time series, is the difference between the mean of the product of  \textbf{x} and \textbf{y} subtracted from the product of the means, the Eq. \ref{appcorr} leads to:
\begin{equation*}
{\small
\textbf{{\sc corr}}(\textbf{x},\textbf{y}) = \frac{\frac{1}{T} \displaystyle\sum_{t=1}^T x_t y_t   -  \mu_{\textbf{x}} \mu_{\textbf{y}}}{\sigma(\textbf{x}) \sigma(\textbf{y})}
}
\end{equation*}
where $\mu_{\textbf{x}}$  and $\mu_{\textbf{y}}$ are the means of \textbf{x} and \textbf{y} respectively, and $\sigma(\textbf{x})$ and $\sigma(\textbf{y})$ are the standard deviations of \textbf{x} and \textbf{y}.

Note that if \textbf{x} and \textbf{y} are standardized\footnote{The datasets from UCR archive are standardized.}, they will each have a mean of 0 and a standard deviation of 1, so:

\begin{equation*}
{\small
\mu_{\textbf{x}} =  \mu_{\textbf{y}} = 0 
}
\end{equation*}
\begin{equation*}
{\small
\sigma(\textbf{x}) =  \sigma(\textbf{y}) = 1
}
\end{equation*}
and the formula reduces to:

\begin{equation}
\label{appcorrreduced}
{\small
\textbf{{\sc corr}}(\textbf{x},\textbf{y}) = \frac{1}{T} \displaystyle\sum_{t=1}^T x_t y_t  
}
\end{equation}

Whereas the Euclidean distance is the sum of squared differences, the Pearson correlation is basically the average product. 
If we expand the Euclidean distance (Ed) formula, we get:

\begin{eqnarray*}
{\small
	d_E(\textbf{x},\textbf{y}) = \sqrt{\displaystyle\sum_{t=1}^T  {(x_{t} - y_{t})}^2}  = \sqrt{\displaystyle\sum_{t=1}^T  {x_{t}}^2 + \displaystyle\sum_{t=1}^T  {y_{t}}^2 - 2  \displaystyle\sum_{t=1}^T  x_{t}  y_{t}}  
}
\end{eqnarray*}
But if \textbf{x} and \textbf{y}  are standardized, the sums $\sum_{t=1}^T  {x_{t}}^2$ and $\sum_{t=1}^T  {y_{t}}^2$ are both equal to T.
That leaves $\sum_{t=1}^T  x_{t}  y_{t}$ as the only non-constant term, just as it was in the reduced formula for the Pearson correlation coefficient (Eq. \ref{appcorrreduced}). 

Thus, for the standardized dataset, One can write the Pearson correlation between two time series in terms of the squared distance between them as:

\begin{equation*}
{\small
\textbf{{\sc corr}}(\textbf{x},\textbf{y}) =  1 - \frac{d_E^2(\textbf{x},\textbf{y})}{2T}. 
}
\end{equation*}
.$\Box$

\end{document}